\documentclass[a4paper,10pt,twocolumn]{article}
\usepackage[utf8]{inputenc}
\usepackage[english]{babel} 
\usepackage{vmargin}
\usepackage{authblk}

\usepackage{graphics}
\usepackage{color}

\usepackage{amssymb}
\usepackage{placeins}
\usepackage{graphicx}			
\usepackage{textcomp}			
\usepackage{amsmath}			

\usepackage{lineno}
\usepackage[hidelinks]{hyperref}
\modulolinenumbers[5]

\usepackage[algoruled]{algorithm2e}
\SetKw{Generate}{Generate}
\SetKw{Evaluate}{Evaluate}
\SetKw{Crossover}{Crossover}
\SetKw{Competition}{Competition}
\SetKw{Reinsertion}{ReInsertion}
\SetKw{Select}{Select}
\SetKw{Mutation}{Mutation}

\setpapersize{A4}
\setmargins{2.5cm} 
{1.5cm}            
{16.5cm}           
{23.42cm}          
{10pt}             
{1cm}              
{0pt}              
{2cm}              

\title{A comparison of different types of Niching Genetic Algorithms for variable selection in solar radiation estimation}
\author[1]{Jorge Bustos}
\author[2]{Victor Adrian Jimenez}
\author[2]{Adrian Will}
\affil[1]{\small Facultad de Ciencias Exactas y Tecnolog\'ia, Universidad Nacional de Tucum\'an}
\affil[2]{\small Grupo de Investigaci\'on en Tecnolog\'ias Inform\'aticas Avanzadas (GITIA), Universidad Tecnol\'ogica Nacional}

\date{}

\begin{document}


\maketitle

\footnotetext[1]{Facultad de Ciencias Exactas y Tecnolog\'ia, U.N.T., Av. Independencia 1800 (4000), San Miguel de Tucum\'an, Tucum\'an, Argentina. E-mail: jebustos@herrera.unt.edu.ar}
\footnotetext[2]{Grupo de Investigaci\'on en Tecnolog\'ias Inform\'aticas Avanzadas, U.T.N., F.R.T., Rivadavia 1050 (4000), San Miguel de Tucum\'an, Tucum\'an, Argentina. E-mail: info@gitia.org}
  
\begin{abstract}

Variable selection problems generally present more than a single solution and, sometimes, it is worth to find as many solutions as possible. The use of Evolutionary Algorithms applied to this kind of problem proves to be one of the best methods to find optimal solutions. Moreover, there are variants designed to find all or almost all local optima, known as Niching Genetic Algorithms (NGA). There are several different NGA methods developed in order to achieve this task.
The present work compares the behavior of eight different niching techniques, applied to a climatic database of four weather stations distributed in Tucuman, Argentina. The goal is to find different sets of input variables that have been used as the input variable by the estimation method. Final results were evaluated based on low estimation error and low dispersion error, as well as a high number of different results and low computational time. A second experiment was carried out to study the capability of the method to identify critical variables. The best results were obtained with Deterministic Crowding. In contrast, Steady State Worst Among Most Similar and Probabilistic Crowding showed good results but longer processing times and less ability to determine the critical factors.

\end{abstract}


\section{Introduction}
\label{Introduction}

The selection and identification of variables that best explain the behavior of a phenomenon is a major issue and difficult to solve, especially when there is a large number of variables. The increasing use of large databases to estimate one or more output variables requires a powerful processing capacity. However, this is no problem for current computers, but the use of traditional machine learning techniques to make these estimates may not be suitable because they generally work properly with small datasets. Therefore, it is very important to reduce the size of the input data set in a system in order to preserve the most important information and apply these techniques correctly.

The Variable Selection or \textit{Feature Subset Selection} is a technique that reduces the size of a dataset by removing the variables with irrelevant, redundant, or noisy data. It is based on the hypothesis that reducing the dimensionality of datasets will reduce estimation errors, model complexity, and computational cost. This technique does not alter the original representation of the variables, contrary to other dimensionality reduction techniques based on data projection (e.g., Principal Component Analysis and Partial Least Squares \cite{Goyal_2010,Kandari_2005}) or data compression using information theory \cite{Knuth_2013}.

In a previous paper \cite{Will_2013} the authors introduced a new methodology for variable selection for solar radiation estimation based on Artificial Intelligence. This work aims to use a Niching Genetic Algorithm (NGA) coupled with linear regression to determine which variables have the greatest influence on the estimation problem. The genetic algorithm system developed can identify variables as critical, redundant, or equivalent to others.
Niching Genetic Algorithm allows the system to produce several different combinations of climatic variables, which render the same estimation error. The preferred algorithm was Deterministic Crowding, which is a very efficient, widely used type of Niching GA \cite{Mahfoud_1995,Yu_2008,Yu_2010,Vollmer_2010}.

Nevertheless, there are several other types of Niching Genetic Algorithms that appear adequate for this task. In this context, it would be a significant contribution to perform a comparison between different NGAs methods, as made by \cite{Singh_2006}. In this work, the authors used three test functions to compare the following Niching methods: Clearing, Clustering, Deterministic Crowding, Probabilistic Crowding, Restricted Tournament Selection, Sharing, Species Conserving GA and Modified Clearing. Their results have shown that restricted tournament selection, deterministic crowding, original clearing, and the proposed modified clearing method did fairly well in problems with reduced search space and the number of local optima. As these variables increase, only the modified clearing method showed good results. In this paper, we implemented eight variants of Niching GA and, in order to compare their performances, we applied them to the problem of variable selection using Linear Regression to estimate global solar radiation for El Colmenar, located in the province of Tucum\'an, Argentina, based on the data from 4 weather stations spread along the North of Argentina \cite{Will_2013}. The comparison was based on estimation errors, efficiency in assessing variable impact, and execution time.

This paper is organized as follows.
In section \ref{Material and Methods}, we present the data used, its organization, and some details about the methodology.
In section \ref{Niching_GA}, the algorithms used in this paper are explained.
In section \ref{Implementacion}, details about the implementation of the algorithms and their use for variable selection in the problem are presented.
In section \ref{Results}, the obtained results are explained.
Finally, in section \ref{Conclusions}, the conclusions and future lines of research are detailed.

\section{Material and Methods}
\label{Material and Methods}

\subsection{Data Collection}
\label{Data_Collection}

Data from 4 different weather stations were used to estimate solar radiation in El Colmenar. These stations are located in the province of Tucuman, Argentina and provide the following variables: day of year, daily average temperature $[^\circ \mbox{C}]$, air humidity [\%], atmospheric pressure [$mbar$], cloudiness [$x/8$] and observed global solar radiation [$MJ/m^2$], to estimate solar radiation in El Colmenar, Tucuman, Argentina, in the period between 01-01-2006 and 12-31-2011.

To estimate global solar radiation on a given day, the following data were used:
\begin{enumerate}
  \setlength{\itemsep}{0pt}
  \setlength{\parskip}{0pt}
 \item Solar global radiation in El Colmenar, for the previous 4 days (endogenous variable).
 \item Temperature, air humidity, atmospheric pressure, cloudiness, and sunshine hours, for all the remaining stations, for the previous 4 days and the day of the estimation (exogenous variables).
 \item Day of Year (Julian Day), only once per prediction day.
\end{enumerate}

This leaves a total of 89 variables, detailed in Table~\ref{tab:Input_variables}.

\begin{table*}[!htb]
 \begin{center}
 \includegraphics[width=0.9\textwidth]{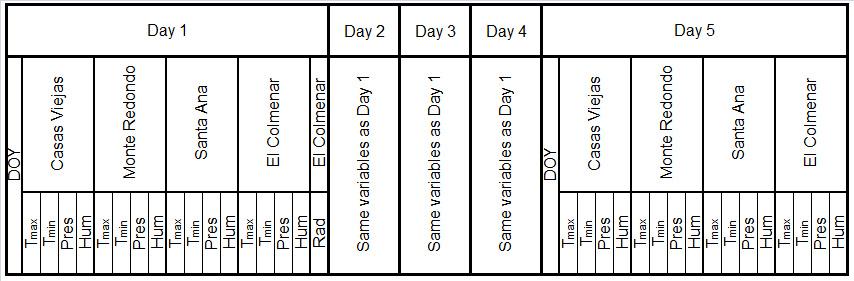}
 \end{center}
 \caption{Input variables summary. The variables for Day 5 contain more recent information (time of estimation); the variables for Day 4 corresponds to one day before; Day 3 for two days before, and so on.}
 \label{tab:Input_variables}
\end{table*}

\subsection{Estimation method for solar radiation}
\label{Explanation Linear_Reg}

In this paper, linear regression is used to estimate solar radiation in El Colmenar, Tucuman, Argentina. Although somewhat better results can be obtained using neural networks, this method has the advantage of low processing time and ease of application. Due to the strong linear correlation present in the data (preliminary tests determined that some of the variables have linear correlation around $0,65$), pseudo-inverse was used in order to avoid undesirable effects due to the ill-condition of the linear systems involved \cite{Elden_2007}.

Since the objective of this work is to compare the performance of different algorithms, we only applied root mean square error ($RMSE$) and the linear correlation coefficient ($R$), since these are enough for that purpose. Moreover, $RMSE$ has the advantage of being expressed in the same units as the output variable, which makes the interpretation of the results easier.


\section{Niching Genetic Algorithms}
\label{Niching_GA}

In a Basic Genetic Algorithm, the objective is to find the global optimum of a problem, or more usually a good approximation to it. Nevertheless, in many cases, it is important to find not only the global optimum but all the local optima. Several modifications of genetic algorithms have been proposed to address the problem of finding all realizations of the local optimums. These are known as \textit{niching algorithms}, based on the biological metaphor of species occupying different ecological niches, where species coexist without competing with each other. In this case, this paradigm allows producing several different combinations of climatic variables, which render the same estimation error. In this paper, we will consider some of the most cited niching methods available in order to evaluate their performance in the problems under consideration.

\subsection{Crowding Methods}

\begin{figure}[!htb]
 \begin{center}
 \includegraphics[width=0.9\columnwidth]{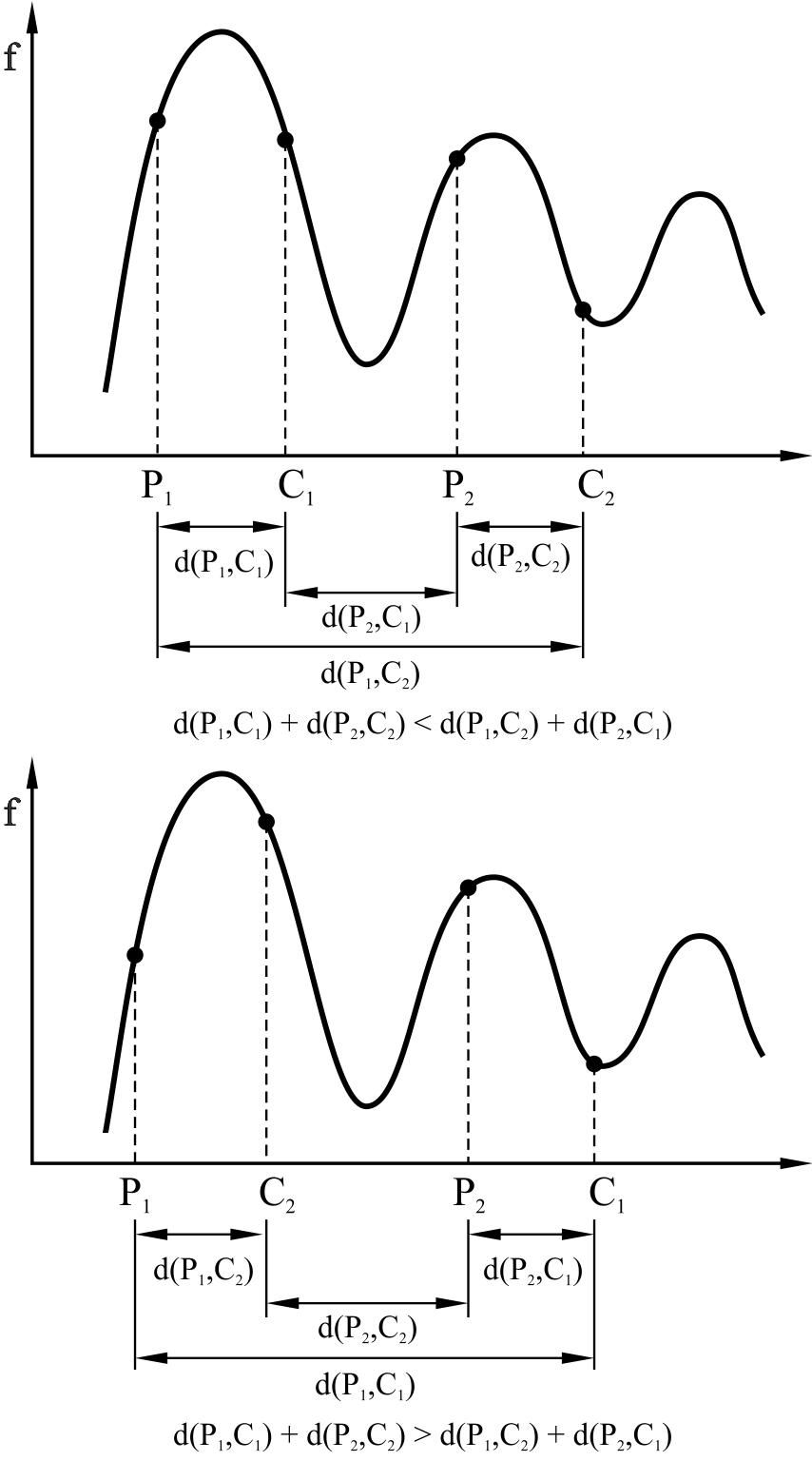}
 \end{center}
 \caption{Pairing phase in Deterministic and Probabilistic Crowding. The vertical axis represents the Fitness Function. $P_1$ and $P_2$ are the parents, while $C_1$ and $C_2$ are the children.}
 \label{fig:esquema_crowding}
\end{figure}

\begin{figure*}[!htb]
 \begin{center}
 \includegraphics[width=0.7\textwidth]{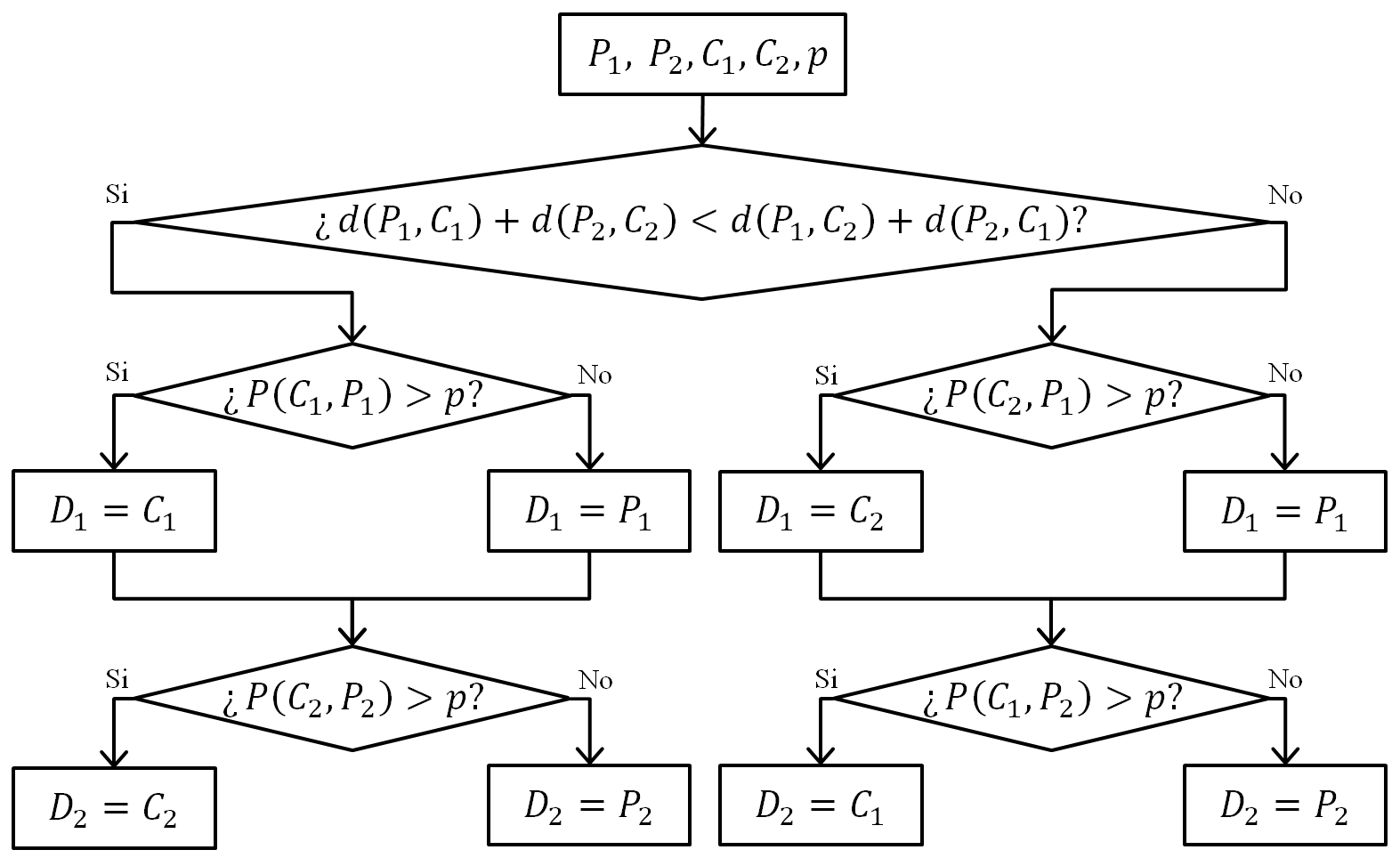}
 \caption{Flowchart of Crowding Algorithms. Parents ($P_1$ and $P_2$), and offspring ($C_1$ and $C_2$) are paired based on a distance $d$. Then it is decided which individual in each pair survives into the next generation, based on a random $p \in (0,1)$ value.}
 \end{center}
 \label{fig:generalized_crowding}
\end{figure*}

Crowding is a set of population variants of Niching GA. The original Crowding method was introduced by De Jong \cite{DeJong_1975} as a way to prevent premature convergence. Its main ingredient is a Population-based GA, in which the selection process has been modified in order to decide which individuals are going to survive into the next generation. It comprises 2 phases: Pairing and Substitution. In the Pairing phase, each individual in the offspring population is paired with an individual in the parents population based on a distance $d$ defined in the search space. In Figure \ref{fig:esquema_crowding}, it is shown schematically the basis of the method, in which the closest individuals are selected for competition. In the Substitution phase, a decision is made as to which of the two individuals (parent or offspring) will survive into the next generation. This decision is made based on a probability function $P(c,p)$, which denotes the probability that child the $c$ replaces the parent $p$ in the population \cite{Galan_2010}. More details on the algorithm crowding shown in Figure \ref{fig:generalized_crowding}.

\subsection{Deterministic and Probabilistic Crowding}
Depending on the function $P$ that determines the substitution, there exist mainly two types of Crowding: Deterministic Crowding (DC) \cite{Mahfoud_1995,Yuan_2002,Dick_2005,Yu_2010} and Probabilistic Crowding (PC) \cite{Mengshoel_1999,Yu_2010}. In Deterministic Crowding, the replacement function $P$ is determined by equation \ref{ec:prob_det_crowding}. The winner of the competition is deterministically taken as the one with greater fitness. This has the advantages of being fast and stable upon repetitions due to the lack of random parts in the algorithm. It can also be noticed that it corresponds to Generalized Crowding for the following function $P$:
\begin{equation}
 P(c,p)=\left\{
 \begin{array}{cl}
 1   & \quad \text{if }f(c) > f(p) \\
 0.5 & \quad \text{if }f(c) = f(p) \\
 0   & \quad \text{if }f(c) > f(p)
 \end{array}
 \right.
 \label{ec:prob_det_crowding}
\end{equation}

On the other hand, Probabilistic Crowding attempts to correct the bias introduced into the algorithm by the deterministic part by introducing a random component into the function $P$, determined by equation \ref{ec:prob_prob_crowding}.
That is, the parent or the children are selected to survive based on random choice. Nevertheless, a random number is necessary for every evaluation of the function $P$, so Probabilistic Crowding is computationally more expensive than Deterministic Crowding. Moreover, this random component makes this version less stable upon repetitions since launching the process twice on the same initial population, and the same parameters will be much less likely to produce similar results.
\begin{equation}
 \begin{aligned}
   P(c,p) &= \frac{f(c)}{f(c)+f(p)} &
 \end{aligned}
 \label{ec:prob_prob_crowding}
\end{equation}

These distance-based methods (as opposed to neigh\-bor\-hood\--based methods) have the important advantage of having \textit{implicit} the size and shape of the neighborhood; that is, it does not have to be explicitly defined.

\subsection{Steady State}

Most of the remaining Niching GA methods are of the Steady State type, usually varying on the reinsertion stage and minor architectural details.
Steady State operates by taking out a small portion of the population in each generation (the ``tournament'') in which, after a few operations, a new candidate is generated and then reinserted into the population.

Steady State methods require a high number of generations to converge, which compensates the relatively little processing time required to process each generation.

\subsubsection{Restricted Tournament Selection (RTS)}

Restricted Tournament Selection is one of the most used Niching methods \cite{Yu_2008,Yu_2010,Vollmer_2010,Lozano_2008}. In this Steady State method, only two solutions are selected randomly from the population in every generation. These two solutions undergo Crossover and Mutation, generating a total of 5 solutions that compete among themselves, generating only 1 candidate for reinsertion, $A$. For the reinsertion operator, $n$ candidates are selected randomly from the population. Then $A'$ is selected as the closest to $A$ with respect to distance $d$, among the $n$ candidates selected. Then $A$ competes with $A'$ and the fittest is reinserted into the population.

\subsubsection{Worst Among Most Similar (WAMS)}

WAMS is a variation of RTS \cite{Cedeno_1999}. In this method, a candidate $A$ for reinsertion is generated the same way as in RTS. Then the population is split randomly into subgroups of $n$ solutions each, and the closest in each group to $A$ is selected for competition. Then the one with the lowest fitness among this group is replaced by the new candidate $A$, even if it has lower fitness than other individuals.
In the variant implemented in this work both, old and new individuals, compete with each other, and the one with better fitness is reinserted.

\subsubsection{Enhanced Crowding (EC)}

This method is similar to WAMS. The difference lies in the reinsertion stage. At this point, the individual with the worst fitness of each group is selected to go into the next group. Among those in this group, the one closest to the candidate competes with the new candidate and the one with better fitness goes back into the population \cite{Cedeno_1994}.

\subsubsection{Fitness sharing (FS)}

Fitness Sharing \cite{Dick_2005,Goldberg_1987,Vollmer_2010,Yu_2010} is based on the principle of distribution of limited resources. Then an overcrowded niche will be penalized, and its population forced to move and populate more vacant niches or even discover new ones. The shared fitness can be calculated as
\begin{equation}
 f_{share} = \frac{f_0}{m_i}
\end{equation}

where $f_0$ is the original fitness, $m_i$ is the number of individuals in the same niche as the solution.

\section{Implementation}
\label{Implementacion}

For comparison purposes, it is necessary to implement the different algorithms as similar as possible, although it is well known that implementation details strongly depend on each algorithm. In order to do this, codification and fitness function are the same for all methods, while operators like selection, crossover, and mutation maintain similarity when possible. The number of generations must change strongly from one method to another in order to adapt to the algorithm, e.g., population methods use much fewer generations than Steady State methods.

\subsection{Codification}
In order to implement Niching GAs, it is necessary to code it properly. For variable selection, individuals can be coded as binary vectors, where a number 1 corresponds to a selected variable, and a zero corresponds to an unselected variable. This kind of binary codification keeps individuals sizes unchanged.

\subsection{GA's fitness function}
\label{Fitness_function}

The fitness function is based on the prediction error found by the linear regression on the training data. A penalization factor was included, favoring data from nearer cities. In particular, the fitness function is based on the root mean square error ($RMSE$) and linear correlation coefficient ($R$) errors of the linear regression estimation over the training data. Also, the penalization takes into account the distance from the weather station to the point of interest (in this case, El Colmenar and the number of variables effectively used by the particular solution being evaluated.

First, a normalized geographic distance is obtained for each weather station, dividing the distance from the weather station $i$ to the station of the variable of interest, $d_i$, by the distance to the farthest station considered $d_{max}$. This provides us with a distance penalization factor for each weather station, between $0$ and $1$. Each climatic variable effectively used by the solution (coordinate equal to 1) is multiplied by the corresponding penalization factor of the corresponding weather station. The sum of all these gives a penalization factor $Pen$:
\begin{equation}
 \begin{aligned}
   Pen(Sol) &= \sum_i \frac{sol(i)\cdot d_i}{d_{max}} &= \sum_{Sol(i) \neq 0} \frac{d_i}{d_{max}}
 \end{aligned}
\end{equation}

where $Pen$ is the factor and $sol$ is the solution. This factor takes into account the number of variables a particular solution uses and from which station it is taken. This strategy favors solutions with fewer variables, and when given two solutions with the same number of variables, it will favor the one that uses variables from the nearer stations.

The fitness function $F$ is then calculated as
\begin{equation}
 \begin{aligned}
  F(Sol) &= \frac{RMSE}{R} \left[1 + 1.5 \cdot Pen(Sol) \right] \\
 \end{aligned}
\end{equation}

\section{Results and Discussion}
\label{Results}
In this work, the objective is twofold: On one hand, to assess the impact of the different algorithms in variable selection for solar radiation estimation, both on data imputation and variable impact analysis. In both cases, the Niching genetic algorithm is used to determine different combinations of variables that produce similar prediction errors. A minimum number of generations is necessary so that the algorithm can locate and establish local optimal points while reducing the estimation error and diminishing the number of variables (see penalty vector \cite{Will_2013}.

Several runs on the different algorithms were conducted, since the different algorithms present very different characteristics, both in the number of individuals and generations. Moreover, the analysis needs to differentiate among crowding and niching, i.e., the number of generations is a compromise between the niching structure (i.e., several local optimal solutions need a minimum of generations to maintain variability) and the prediction error (that needs a certain number of generations in order to produce good solutions). If enough generations go by, many methods tend to present \textit{Crowding or premature convergence}, which is desirable for the impact analysis of the variables, but not for the imputation of the missing data.

\subsection{Analysis of Niching Genetic Algorithms for Variable Selection}

\begin{figure*}[!htb]
 \begin{center}
 \includegraphics[width=0.73\textwidth]{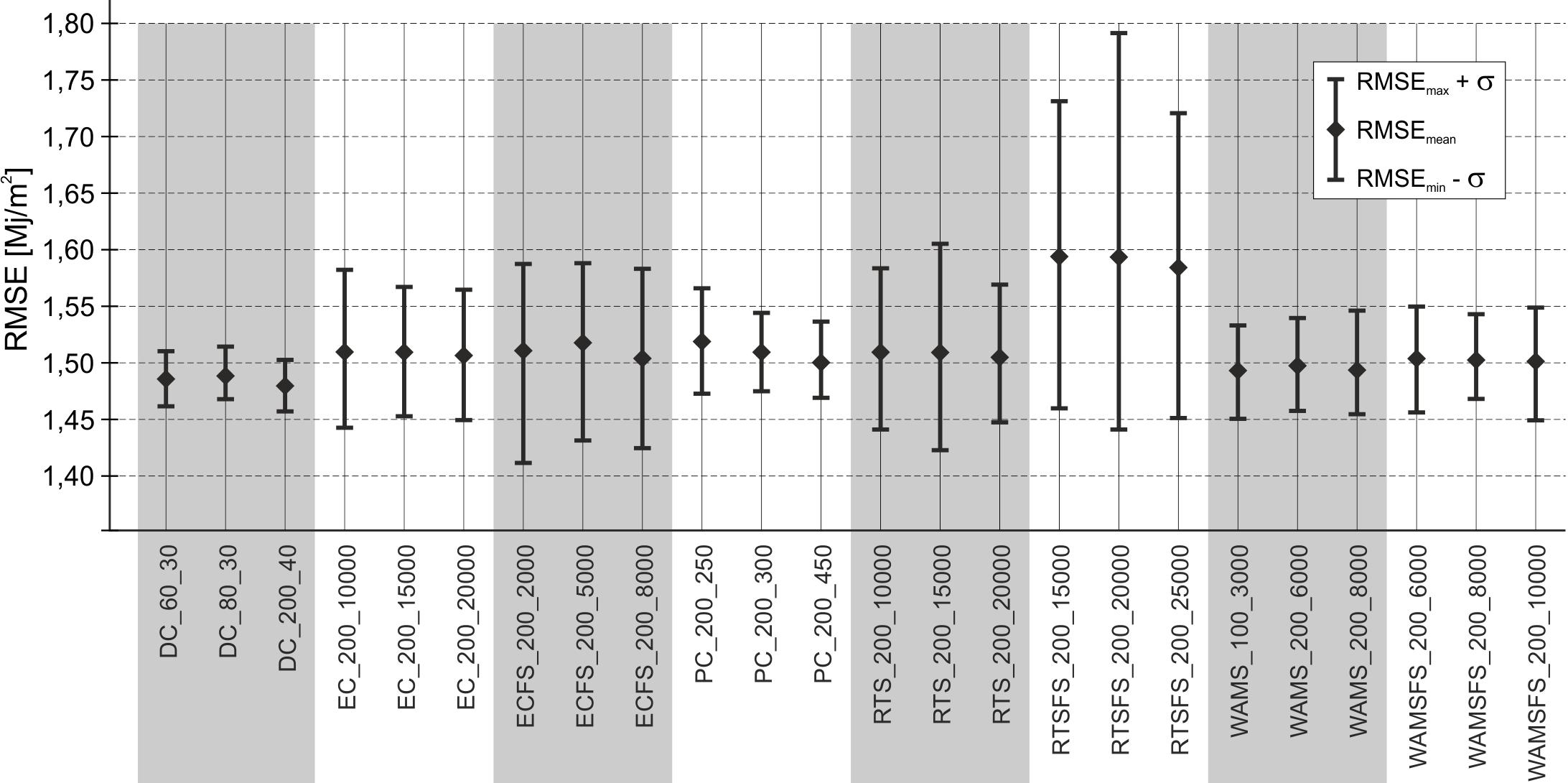}
 \end{center}
 \caption{Estimation Error (Average and Dispersion) - All Niching Methods.}
 \label{fig:Errors_All_Alg}
\end{figure*}

Two sets of tests were conducted. In the first round, all eight algorithms were tested in order to determine the point of compromise between the stable low prediction error and the niching structure described above. In this first round, results showed that most of the algorithms, excluding RTS with Fitness Sharing, have an average estimation error that stayed within $1,47$ and $1,52$ [$MJ/m^2$], with dispersion ranges between $0,01$ up to $0,18$ [$MJ/m^2$], as shown in Figure \ref{fig:Errors_All_Alg}. In some methods, errors and STD result too large to be practical, so the estimation is not stable, and many solutions are useless.

As for the variability analysis, final solutions of each method were considered. In order to discriminate variable importance, they can be classified as follows:

{\bf Critical variables}: Variables that were selected in more than or equal to $95\%$ of the final population.

{\bf Non-critical variables}: Variables that were selected in more than or equal to $5\%$ and less than $95\%$ of the final population.

{\bf Irrelevant variables}: Variables that were selected in less than $5\%$ of the final population.

\begin{figure*}[!htb]
 \begin{center}
 \includegraphics[width=0.73\textwidth]{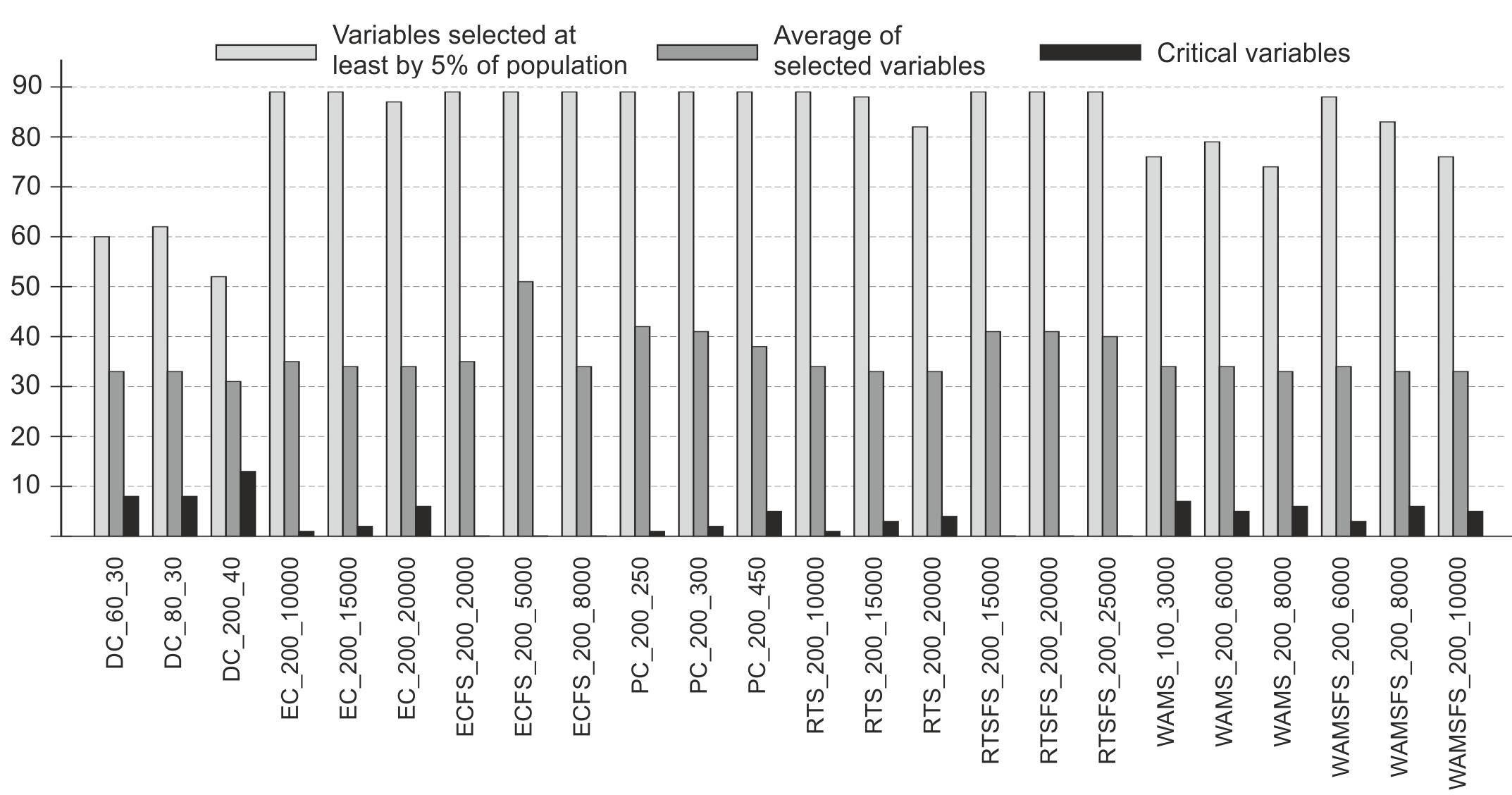}
 \end{center}
 \caption{Variables Selected - Critical, non-critical, and average - All Niching Methods.}
 \label{fig:Variables_Selected_All_Alg}
\end{figure*}

Also, the average number of selected variables was considered in each method, as shown in Figure \ref{fig:Variables_Selected_All_Alg}

\begin{figure*}[!htb]
 \begin{center}
 \includegraphics[width=0.69\textwidth]{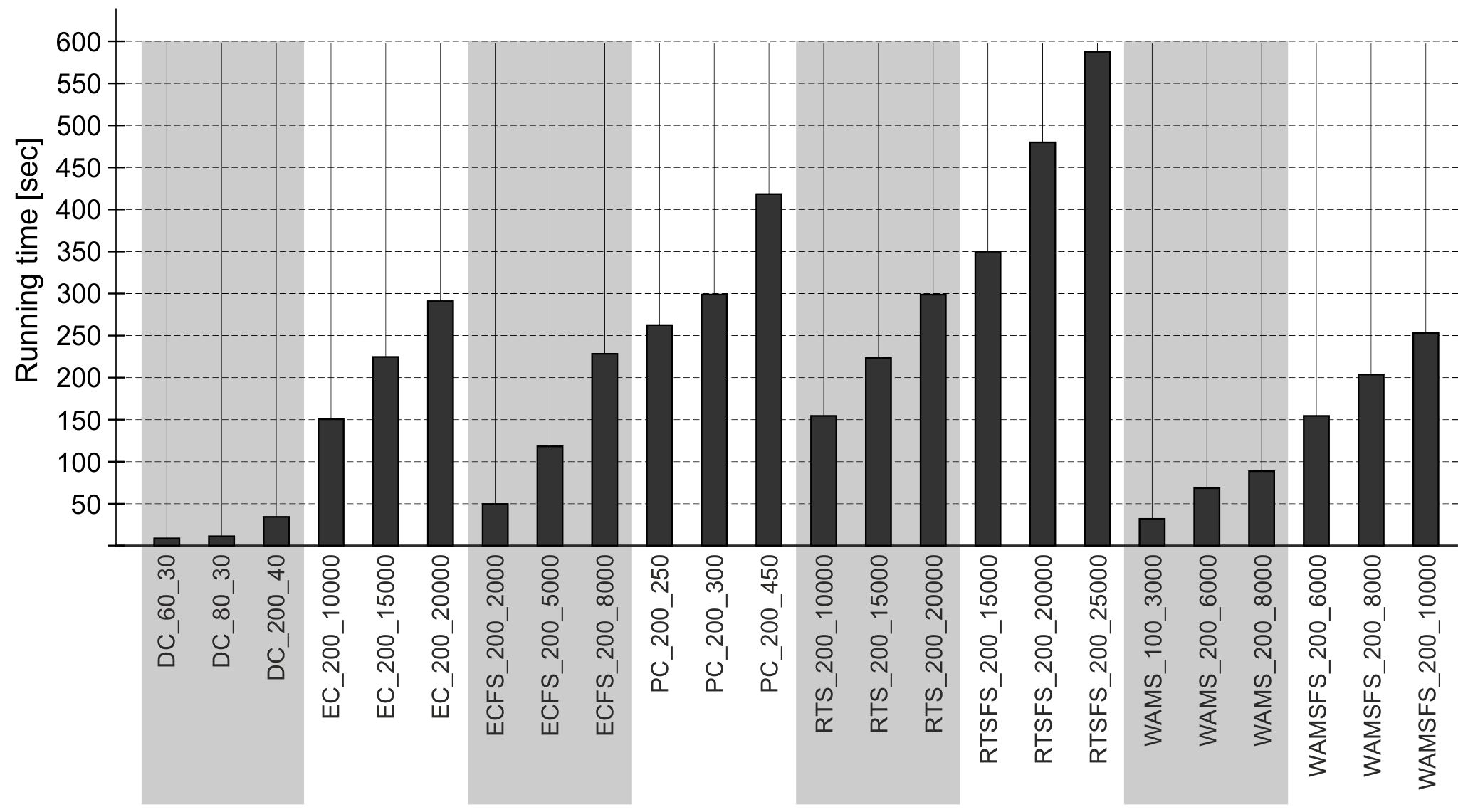}
 \end{center}
 \caption{Running times - All Niching Methods.}
 \label{fig:Running_Times_All_Alg}
\end{figure*}

\begin{figure}[!htb]
 \begin{center}
 \includegraphics[width=1\columnwidth]{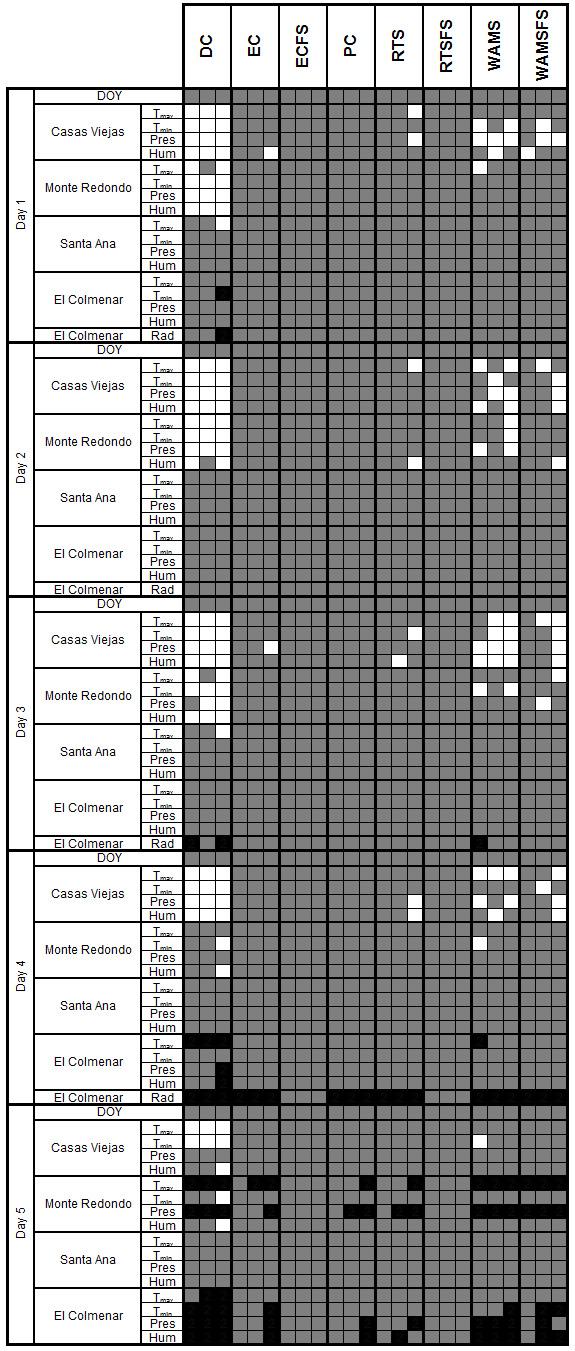}
 \end{center}
 \caption{Selected variables - Black=Critical, Grey=Non critical, White=Not selected - All Niching Methods.}
 \label{fig:Variable_map_All_Alg}
\end{figure}

In Figure \ref{fig:Variable_map_All_Alg}, selected critical and non-critical variables are shown as a colored map. It can be seen that there are different regions of critical, non-critical, and unselected variables for each method.

Running times were also compared (Figure \ref{fig:Running_Times_All_Alg}, showing big differences between methods, from less than $100$ $[secs]$ for DC and WAMS, to an average of $250$ $[secs]$ for EC, ECFS, PC, RTS and WAMSFS and even almost $600$ $[secs]$ for RTSFS.

The second round of tests was conducted in order to perform a variable impact analysis continuing the work of Will et al. \cite{Will_2013}, but only on the three algorithms that appeared as more suitable (Deterministic Crowding, Probabilistic Crowding, and WAMS). In all three cases, the tests were continued up to the point of crowding and where almost all solutions are exactly the same.

In this first set of tests, the objective is to minimize the estimation error while maintaining the niching structure. So, an analysis of critical and redundant variables was performed for each of the eight algorithms (see Figure \ref{fig:Variables_Selected_All_Alg}. In that figure, it can be seen that variants of Generalized Crowding (GC, DC and PC) require fewer generations in order to converge and the algorithms based on a Steady State Architecture (RTS, EC, WAMS and their versions with Fitness Sharing) can require up to 20.000 generations to achieve similar results, but sometimes with larger errors (see Figure \ref{fig:Errors_All_Alg}.
The complete list of variables chosen as critical or non-critical can be found in Figure \ref{fig:Variable_map_All_Alg}.

We can see that most critical variables appear on the same day as the prediction and that the prediction is not stable between different algorithms at this point. This was to be expected since, as shown in \cite{Will_2013}, even repetitions of the same algorithm do not produce the same set of critical variables unless crowding is reached. Also, it is noticeable that ECFS and RTSFS produce no critical variables at this point.

This analysis is completed with a running time comparison for all algorithms. Figure \ref{fig:Running_Times_All_Alg} reports the average of 5 runs of each algorithm and combination, in a PC with a 3,1 GHz Intel Core i5 3450 processor, 8 GB ram DDR3 1600 Mhz and a 1 TB HDD (no GPU or Hardware acceleration used).
In this test, it can be easily appreciated that DC and WAMS produce the lowest running times and Fitness Sharing adds a sizeable amount of processing time to WAMS and RTS, while decreasing the running time of EC.

\subsection{Analysis of the best three algorithms}
To continue the analysis, in the second set of tests, the best three algorithms (Deterministic Crowding, Probabilistic Crowding, and Worst Among Most Similar) were selected on the basis of low estimation error and low error dispersion, as seen in Figure \ref{fig:Errors_All_Alg}. These algorithms were selected as more appropriate for a variable impact analysis. This analysis would consist of continuing the process on each algorithm up until the point of Crowding, by increasing the size of the population and the number of generations.

\begin{figure}[!htb]
 \begin{center}
 \includegraphics[width=1.0\columnwidth]{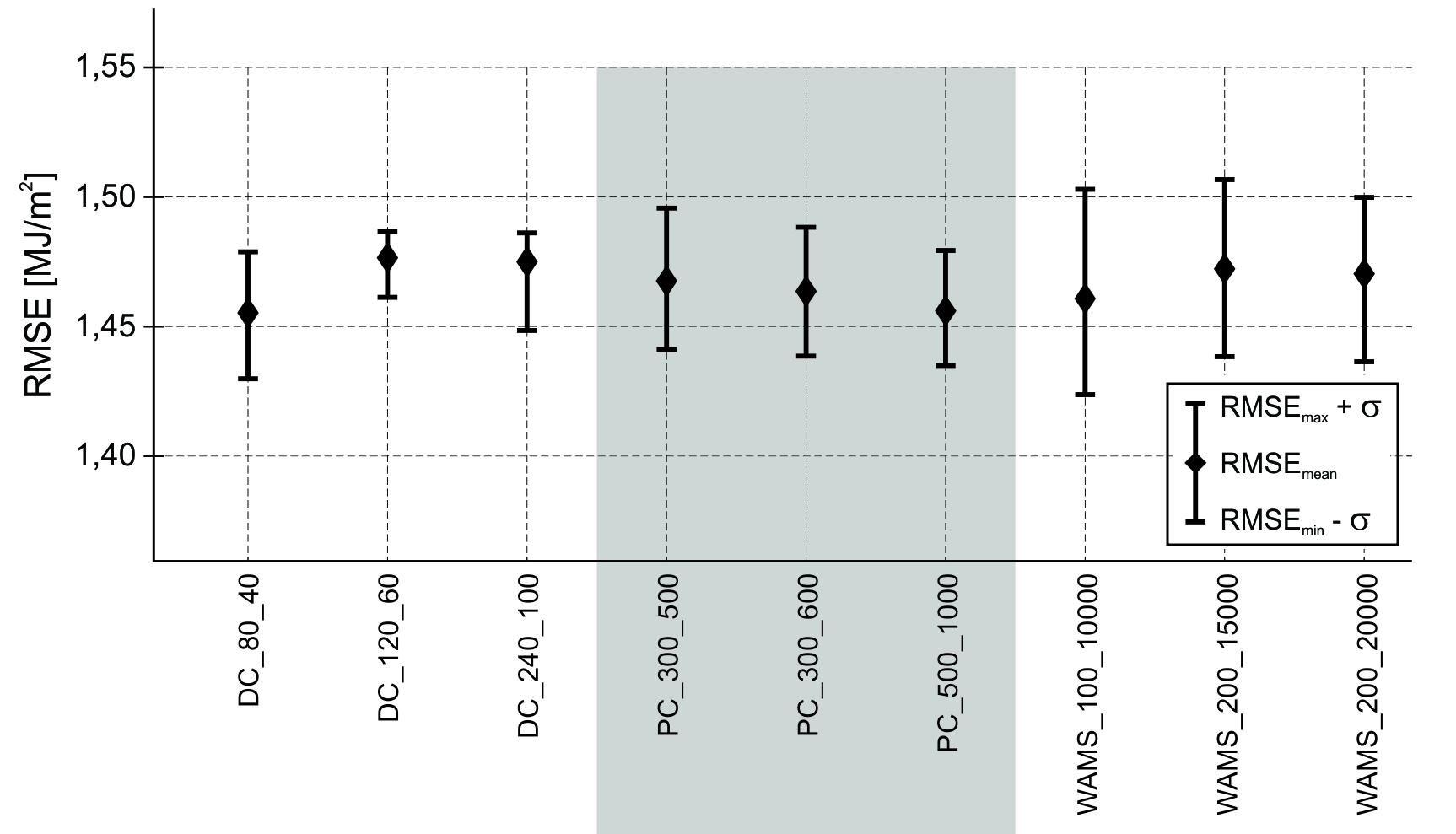}
 \end{center}
 \caption{Estimation Error (Average and Dispersion) - Variable Impact Analysis.}
 \label{fig:Errors_VarCrit}
\end{figure}

\begin{figure}[!htb]
 \begin{center}
 \includegraphics[width=1.0\columnwidth]{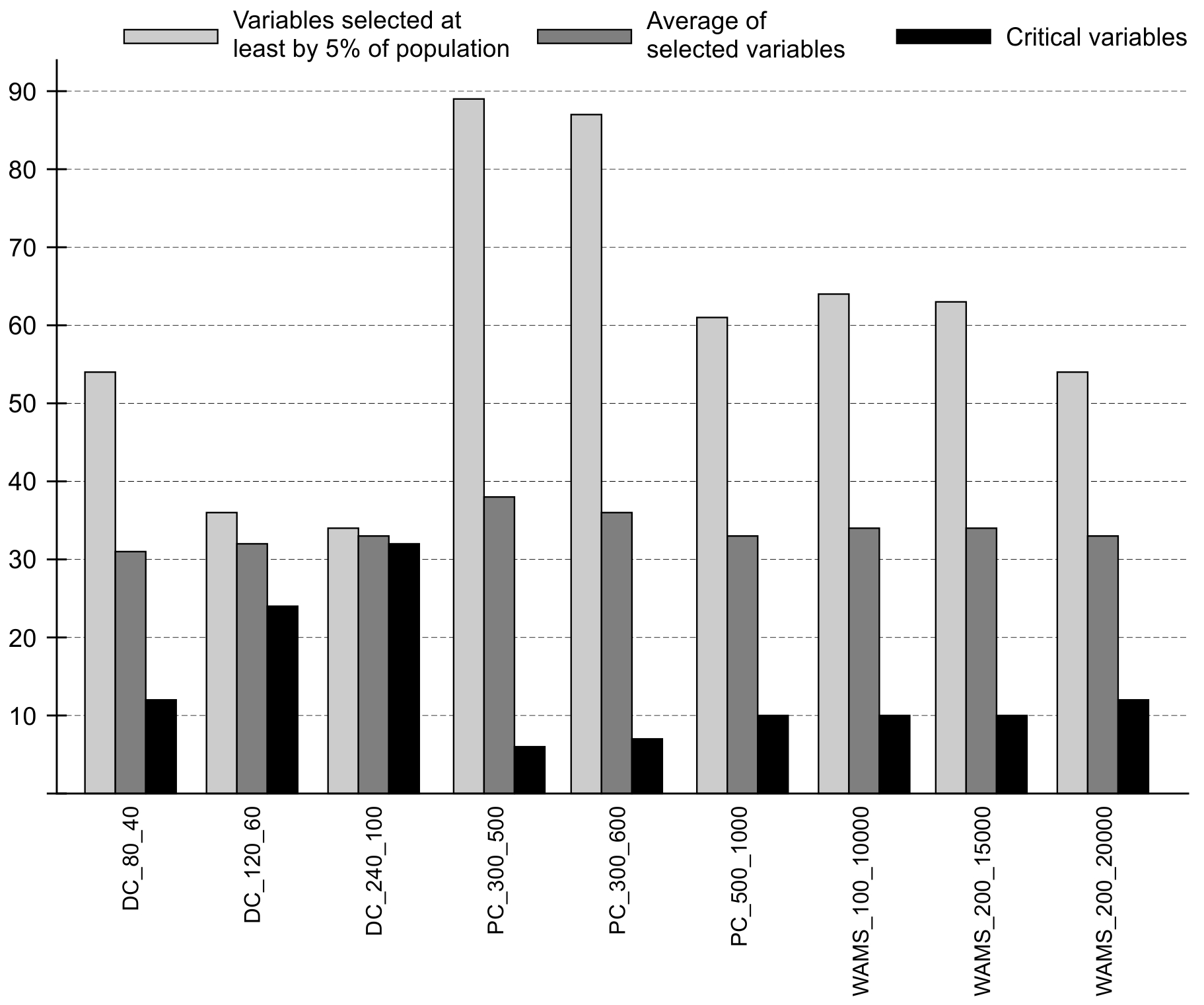}
 \end{center}
 \caption{Variables Selected - Critical, non-critical, and average - Variable Impact Analysis.}
 \label{fig:Variables_Selected_VarCrit}
\end{figure}

\begin{figure}[!htb]
 \begin{center}
 \includegraphics[width=0.8\columnwidth]{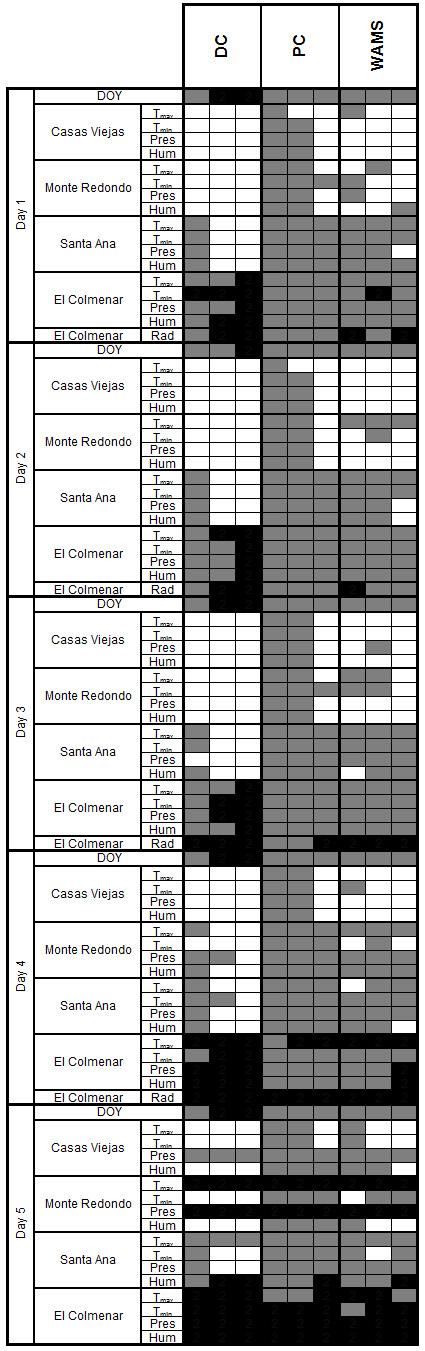}
 \end{center}
 \caption{Selected variables - Black=Critical, Grey=Non critical, White=Not selected - Variable Impact Analysis.}
 \label{fig:Variable_map_VarCrit}
\end{figure}

As generations increase, it can be observed that the errors and the niching structure decrease. At the same time, the number of critical variables also increases, especially in Deterministic Crowding, as shown in figures \ref{fig:Errors_VarCrit}, \ref{fig:Variables_Selected_VarCrit} and \ref{fig:Variable_map_VarCrit}.

\section{Conclusions}
\label{Conclusions}
In this paper, a comparison of different Niching Genetic Algorithms applied to the climatic variable selection problem was presented. Eight different types of Niching GA were implemented and tested on the same data using the same fitness function, selection, and crossover operators. Also, the results were compared on the basis of estimation error and variability, computational performance, ability to classify the variables correctly in critical, non-critical and irrelevant.
Results showed that different algorithms present very different behavior, although the search space and the fitness function are the same. Some algorithms show strictly niching behavior, maintaining the niching structure and variability of the solution at the expense of a large estimation error. The rest of the algorithms find similar estimation errors while maintaining a good niching structure and mainly differ in time complexity considerations.

Deterministic Crowding, Probabilistic Crowding and Worst Among Most Similar algorithms were taken further of the crowding point, in order to assess their ability to classify variables according to their impact on the output variable. Results showed that all three methods present good estimation errors (between 1,45 and 1,47 [$MJ/m^2$]). Deterministic Crowding found the highest number of critical variables (in the most crowded results, 32 variables out of a total of 33 variables selected), Worst Among Most Similar and Probabilistic Crowding found a much lesser number of critical variables (12 and 10 critical variables respectively, out of 33 variables selected, in the most crowded solutions). Computation times showed that Deterministic Crowding obtained the best results in less time than the other methods (85 seconds against 220 seconds for WAMS and 2073 seconds for PC). Furthermore, Probabilistic Crowding and WAMS did not get to a real crowding point. This is probably due to the fact that they both maintain the niching structure for a much larger number of generations, at the price of maintaining large estimation error dispersion (keeping bad solutions alive produces this effect).
Analyzing final results in crowding points (see Figure \ref{fig:Variable_map_VarCrit}, it can be noticed that most selected variables belong to El Colmenar, the same weather station as the estimated variable, in all five days considered. Other variables selected are Day of Year in all five days and air humidity of Santa Ana station on day 5.

Depending then on the task at hand, Deterministic Crowding is one of the best algorithms in all categories, being slightly outperformed by WAMS for the task of niching (it reaches similar niching structure and estimation errors in slightly less processing time).

\section*{Acknowledgments}
This work was partially supported by grant PID-UTN 3870 for the research project ``Intelligent Data Analysis Applied to Energy Management and Optimization''.
We would like to thank the Agrometeorological Section of the Estacion Experimental Obispo Colombres (E.E.A.O.C.), Tucuman, Argentina, for providing the main part of the data used in this work.


\bibliographystyle{plain}
\bibliography{Bustos_et_Al}

\begin{thebibliography}{10}

\bibitem{Kandari_2005}
Noriah~M. Al-Kandari and Ian~T. Jolliffe.
\newblock Variable selection and interpretation in correlation principal
  components.
\newblock {\em Environmetrics}, 16(6):659--672, 2005.

\bibitem{Cedeno_1999}
Walter Cede\~no and V.~Rao Vemuri.
\newblock Analysis of speciation and niching in the multi-niche crowding ga.
\newblock {\em Theoretical Computer Science}, 229:177--197, 1999.

\bibitem{Cedeno_1994}
Walter Cede\~no, V.~Rao Vemuri, and Tom Slezak.
\newblock Multiniche crowding in genetic algorithms and its application to the
  assembly of dna restriction-fragments.
\newblock {\em Evol. Comput.}, 2(4):321--345, December 1994.

\bibitem{DeJong_1975}
Kenneth~A. De~Jong.
\newblock {\em {Analysis of the behavior of a class of genetic adaptive
  systems}}.
\newblock PhD thesis, University of Michigan, 1975.

\bibitem{Dick_2005}
G.~Dick.
\newblock A comparison of localized and global niching methods.
\newblock In {\em Annual Colloquium of the Spatial Information Research Centre
  (SIRC)}, pages 91--101, New Zealand, 2005.

\bibitem{Elden_2007}
Lars Eld\'en.
\newblock {\em Matrix Methods in Data Mining and Pattern Recognition}.
\newblock Society for Industrial and Applied Mathematics, Philadelphia, PA,
  USA, 2007.

\bibitem{Galan_2010}
Severino~F. Galan and Ole~J. Mengshoel.
\newblock Generalized crowding for genetic algorithms.
\newblock In {\em Proceedings of the 12th annual conference on Genetic and
  evolutionary computation}, GECCO '10, pages 775--782, New York, NY, USA,
  2010. ACM.

\bibitem{Goldberg_1987}
D.~Goldberg and J.~Richardson.
\newblock Genetic algorithms with sharing for multi-modal function
  optimization.
\newblock In {\em Genetic Algorithms and Their Applications: Proceedings of the
  Second International Conference on Genetic Algorithms}, pages 44--49. ICGA,
  1987.

\bibitem{Goyal_2010}
M.~Goyal and C.~Prasad~Ojha.
\newblock Application of pls-regression as downscaling tool for pichola lake
  basin in india.
\newblock {\em Intl. Jour. of Geosciences}, 1:51--57, 2010.

\bibitem{Knuth_2013}
Kevin~H Knuth, Anthony Gotera, Charles~T Curry, Karen~A Huyser, Kevin~R
  Wheeler, and William~B Rossow.
\newblock Revealing relationships among relevant climate variables with
  information theory.
\newblock {\em arXiv preprint arXiv:1311.4632}, 2013.

\bibitem{Lozano_2008}
Manuel Lozano, Francisco Herrera, and Jos{\'{e}}~Ram{\'{o}}n Cano.
\newblock Replacement strategies to preserve useful diversity in steady-state
  genetic algorithms.
\newblock {\em Information Sciences}, 178(23):4421 -- 4433, 2008.
\newblock Including Special Section: Genetic and Evolutionary Computing.

\bibitem{Mahfoud_1995}
S.~W. Mahfoud.
\newblock {\em Niching Methods for Genetic Algorithms}.
\newblock PhD thesis, Illinois Genetic Algorithms Laboratory (IlliGAL), 1995.

\bibitem{Mengshoel_1999}
O.~J. Mengshoel and D.~E. Goldberg.
\newblock Probabilistic crowding: Deterministic crowding with probabilistic
  replacement.
\newblock In {\em Proceedings of the Genetic and Evolutionary Computation
  Conference}, pages 409--416. GECCO, 1999.

\bibitem{Singh_2006}
Gulshan Singh and Kalyanmoy Deb, Dr.
\newblock Comparison of multi-modal optimization algorithms based on
  evolutionary algorithms.
\newblock In {\em Proceedings of the 8th Annual Conference on Genetic and
  Evolutionary Computation}, GECCO '06, pages 1305--1312, New York, NY, USA,
  2006. ACM.

\bibitem{Vollmer_2010}
D.T. Vollmer, T.~Soule, and M.~Manic.
\newblock A distance measure comparison to improve crowding in multi-modal
  optimization problems.
\newblock In {\em Resilient Control Systems (ISRCS), 2010 3rd International
  Symposium on}, pages 31--36, 2010.

\bibitem{Will_2013}
A.~Will, J.~Bustos, M.~Bocco, J.~Gotay, and C.~Lamelas.
\newblock On the use of niching genetic algorithms for variable selection in
  solar radiation estimation.
\newblock {\em Renewable Energy}, 50(0):168 -- 176, 2013.

\bibitem{Yu_2008}
L.~Yu and P.N. Suganthan.
\newblock Empirical comparison of niching methods on hybrid composition
  functions.
\newblock In {\em Evolutionary Computation, 2008. CEC 2008. (IEEE World
  Congress on Computational Intelligence). IEEE Congress on}, pages 2194--2201,
  2008.

\bibitem{Yu_2010}
X.~Yu and M.~Gen.
\newblock {\em Introduction to Evolutionary Algorithms}.
\newblock Decision Engineering. Springer, 2010.

\bibitem{Yuan_2002}
B.~Yuan.
\newblock Deterministic crowding, recombination, and self-similarity.
\newblock In {\em Proceedings of the 2002 CEC Congress on Evolutionary
  Computation}, volume~2, pages 1516--1521. IEEE Press, 2002.

\end{thebibliography}

\end{document}